\documentclass[runningheads]{llncs}
\usepackage[T1]{fontenc}
\usepackage{graphicx}
\usepackage{cite}
\usepackage{amssymb, amsfonts}
\usepackage{algorithmic}
\usepackage{textcomp}
\usepackage{ctable}
\usepackage{multirow}
\begin{document}
\title{AM-SORT: Adaptable Motion Predictor with Historical Trajectory Embedding for Multi-Object Tracking
}
\titlerunning{AM-SORT: AM Predictor with Historical Trajectory Embedding for MOT}
%
\author{Vitaliy Kim\orcidID{0009-0000-2031-7599} \and
Gunho Jung\orcidID{0000-0002-1143-9663} \and
Seong-Whan Lee\orcidID{0000-0002-6249-4996}
}
\authorrunning{V. Kim et al.}
%
\institute{Department of Artificial Intelligence, Korea University, Seoul, Republic of Korea \\
\email{\{vitaliy, gh\_jung, sw.lee\}@korea.ac.kr}}
\maketitle              
\begin{abstract}

    Many multi-object tracking (MOT) approaches, which employ the Kalman Filter as a motion predictor, assume constant velocity and Gaussian-distributed filtering noises.
    These assumptions render the Kalman Filter-based trackers effective in linear motion scenarios. 
    However, these linear assumptions serve as a key limitation when estimating future object locations within scenarios involving non-linear motion and occlusions.
    To address this issue, we propose a motion-based MOT approach with an adaptable motion predictor, called AM-SORT, which adapts to estimate non-linear uncertainties.
    AM-SORT is a novel extension of the SORT-series trackers that supersedes the Kalman Filter with the transformer architecture as a motion predictor.
    We introduce a historical trajectory embedding that empowers the transformer to extract spatio-temporal features from a sequence of bounding boxes. 
    AM-SORT achieves competitive performance compared to state-of-the-art trackers on DanceTrack, with 56.3 IDF1 and 55.6 HOTA. 
    We conduct extensive experiments to demonstrate the effectiveness of our method in predicting non-linear movement under occlusions.

\end{abstract}
%
%
%
\section{Introduction}\label{sec:intro}

    Motion-based multi-object tracking (MOT) approaches \cite{bewley2016simple, lee2001automatic, zhou2020tracking, tokmakov2021learning, wang2021track, cao2023observation} utilize a motion predictor to extract spatio-temporal patterns and estimate object motion in future frames for subsequent object association. 
    The original Kalman Filter \cite{kalman1960contributions} is widely employed as a motion predictor, which operates under assumptions of constant velocity and Gaussian-distributed noises in the prediction and filtering stages, respectively \cite{bewley2016simple}.
    Constant velocity postulates that object speed and direction remain consistent over a short period, and Gaussian distributions assume constant error variance in both estimations and detections. 
    While these assumptions result in resource efficiency for the Kalman Filter by simplifying mathematical modeling, they are only valid for a specific scenario where the object displacement remains linear or consistently small at each time step \cite{wang2021track}.
    Due to the neglect of scenarios with non-linear motion and occlusions, the Kalman Filter inaccurately estimates object locations in complex situations.
    
    \begin{figure}[t]
        \centering
        \includegraphics[scale=0.07]{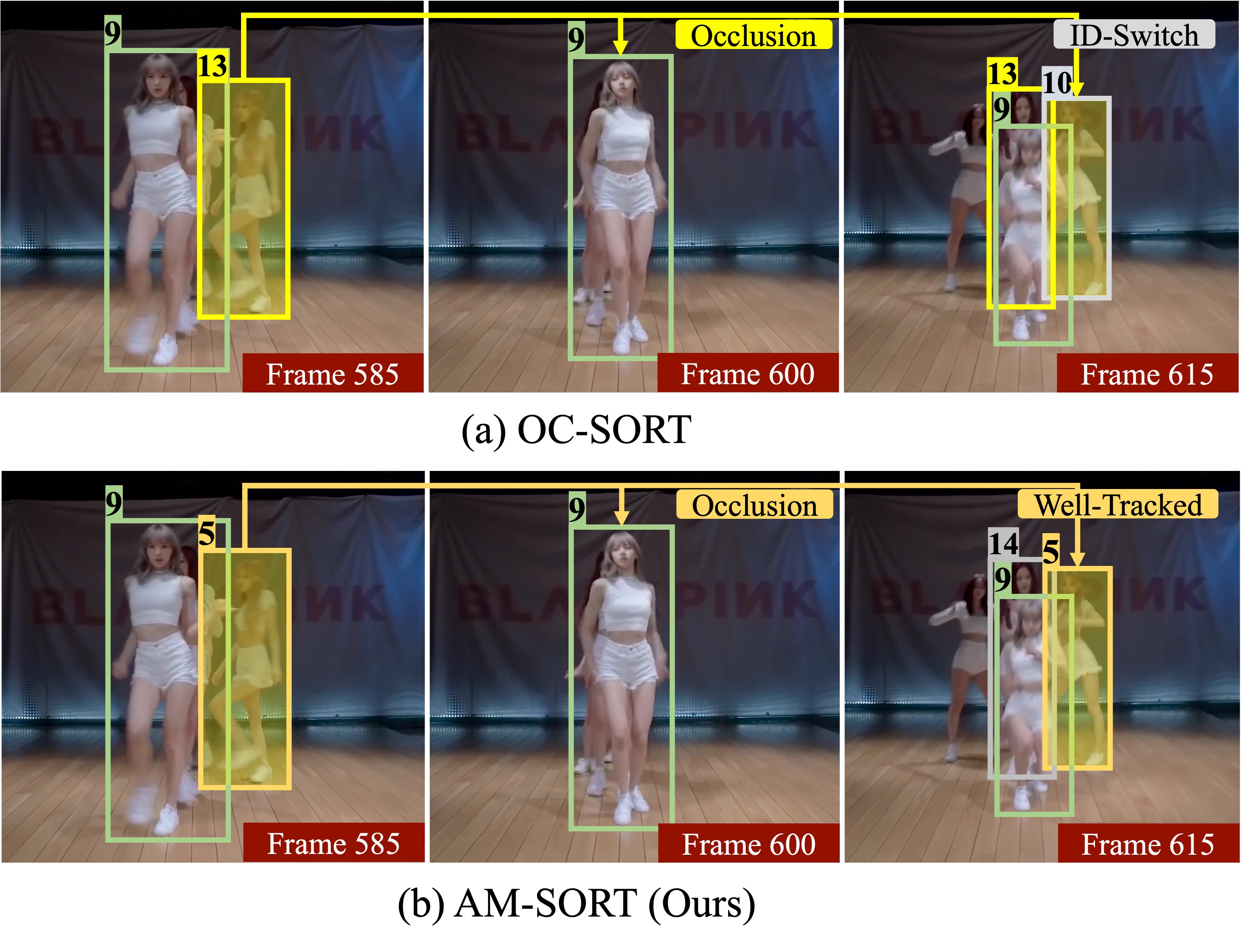}
        \vspace{-0.4cm}
        \caption{Results on \textit{dancetrack0004} sequence from DanceTrack for (a) OC-SORT and (b) AM-SORT (Ours). The object, marked in yellow, moves to the left and becomes occluded in the middle frame. Then, the yellow object changes the movement direction to the right after occlusion, and OC-SORT does not capture this sudden directional shift, causing an ID-switch from 13 to 10.
        \vspace{-0.4cm}
        }
        \label{fig:issue}
    \end{figure}

    To address the limitations of the original Kalman Filter, alternative estimation algorithms were proposed, such as Extended Kalman Filter (EKF) \cite{smith1962application} and Unscented Kalman Filter (UKF) \cite{julier1997new}. 
    EKF linearizes object motion modeling, and UKF estimates non-linear transformations by employing the first and third-order Taylor series expansions, respectively. 
    However, both methods are still conditioned on linear approximations for non-linear systems and assume Gaussian-distributed noises.
    On the other hand, particle filters \cite{gustafsson2002particle} avoid linearization by utilizing a set of discrete particles to handle non-linearity and non-Gaussian noises, yet require expensive computational resources.
    Recent OC-SORT \cite{cao2023observation} improved the original Kalman Filter by placing a greater emphasis on observations rather than estimations to reduce noises in motion prediction.
    While this approach allows for tracking objects with linear motion during occlusions, OC-SORT still faces challenges with non-linear motion.
    When the lack of observations occurs caused by non-linear motion or occlusions, OC-SORT relies on its linear estimations, formulated upon the linear assumptions inherent to the Kalman Filter.
    Consequently, this linear assumption-based modeling accumulates errors in motion prediction leading to significant trajectory deviations.

    \begin{figure}[t]
        \centering
        \includegraphics[scale=0.113]{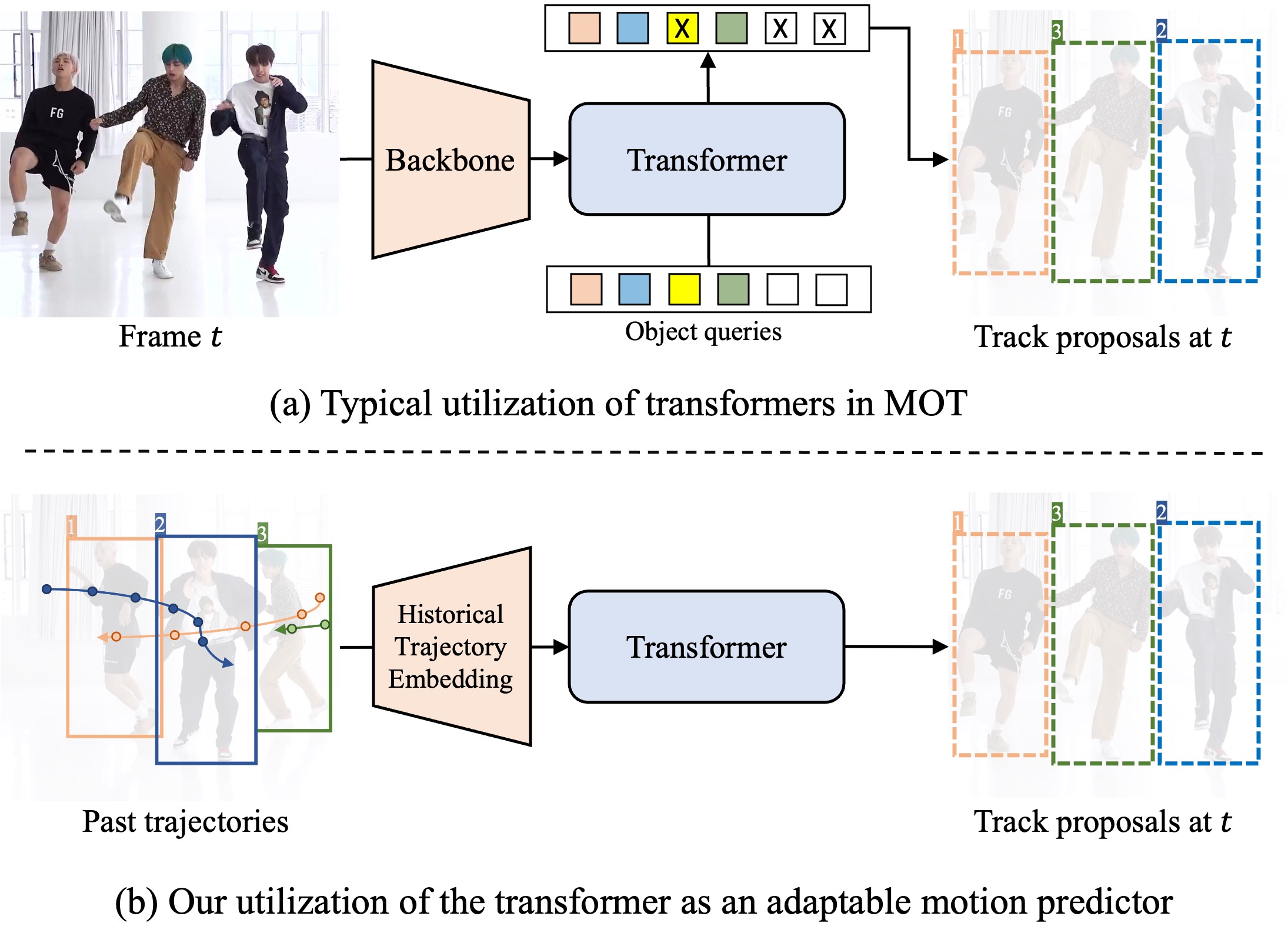}
        \vspace{-0.4cm}
        \caption{Comparison of (a) conventional transformer-based MOT and (b) our frameworks. The key difference lies in the input feature level: typical transformer-based approaches take frames as input and primarily utilize appearance information, whereas AM-SORT processes bounding boxes and solely relies on motion information.
        \vspace{-0.4cm}
        }
        \label{fig:rel_work}
    \end{figure}
        
    We argue that the linear assumptions inherent to the Kalman Filter lead to inaccurate motion estimations and false identity matches when objects involve non-linear uncertainties characterized by sudden speed changes, directional shifts and occlusions.
    Due to these assumptions, the accumulated errors in motion estimations restrict the Kalman Filter-based approaches in handling non-linear uncertainties.
    Fig. \ref{fig:issue} shows tracking results in a non-linear motion scenario under occlusion using (a) OC-SORT and (b) our AM-SORT.
    As illustrated in Fig. \ref{fig:issue}(a), the identity switch occurs for the yellow object after an occlusion event.
    The linear motion assumptions in the Kalman Filter cause directional errors in motion estimations that the yellow object continues moving to the left.
    As a result, the Kalman Filter relies on these linear estimations with accumulated directional errors, failing to predict the directional shift to the right.

    In this paper, we propose an adaptable motion predictor with historical trajectory embedding for MOT that addresses the limitations of the linear assumptions inherent to the Kalman Filter. 
    The adaptation ability releases the motion predictor from the constraints of linear assumptions, allowing it to estimate uncertainties related to non-linear motion. 
    Inspired by transformer architectures \cite{vaswani2017attention, dosovitskiy2020image, giuliari2021transformer}, known for their ability to capture complex dependencies in sequence data, we explore the utilizing of a transformer encoder as an adaptable motion predictor.
    In contrast to conventional transformer-based MOT approaches, we leverage the transformer to encode only motion information without visual features for object association, as shown in Fig. \ref{fig:rel_work}.
    Utilizing bounding boxes as input features provides a limited object representation compared to appearance information but significantly reduces computational complexity. 
    To maintain simplicity and resource efficiency comparable to the Kalman Filter, we focus on the transformer encoder to learn object discrimination features exclusively from object trajectories.
    
    Furthermore, our adaptable motion predictor derives benefits from analyzing and observing longer object trajectories compared to the Kalman Filter, which predicts object motion solely based on estimations from the previous time step.
    To enhance the representation of long object trajectories, we present historical trajectory embedding that encodes the spatio-temporal information from the sequence of bounding boxes. 
    Consequently, we concatenate the embedded bounding boxes with a prediction token that functions as an embedded bounding box of the current frame.
    The encoder extracts the spatio-temporal features from the historical trajectory embedding, enabling the prediction token to estimate the bounding box in the current frame. 
    Notably, AM-SORT utilizes sequences of bounding boxes as input, omitting the visual features of objects, which enables the model to process with low computational cost.

    Our contributions are summarized as follows: 
    \begin{itemize}
        \item We propose a novel SORT-series tracker with an adaptable motion predictor, called AM-SORT, which provides non-linear motion estimations without linear assumptions;
        \item We introduce historical trajectory embedding to effectively capture motion features from a sequence of bounding boxes;
        \item  The qualitative results show that AM-SORT accurately predicts the non-linear changes in object motion, demonstrating its competitiveness with state-of-the-art approaches.
    \end{itemize}

\section{Related Work}\label{sec:related_work}

    \subsection{Motion-Based Methods in Multi-Object Tracking}\label{ssec:motion_rel}

        DanceTrack \cite{sun2022dancetrack} reveals the limitations of appearance-based MOT methods in distinguishing objects that share highly similar visual features. 
        This motivates the development of motion-based and hybrid methods that leverage both appearance and motion information. 
        \cite{bewley2016simple, zhou2020tracking, ahmad2006human, tokmakov2021learning, wang2021track, cao2023observation} propose trackers that solely employ motion features without appearance information.
        CenterTrack \cite{zhou2020tracking} introduces an efficient tracker that represents each object as a single point and predicts their associations with minimal input as detections from a pair of frames.
        PermaTrack \cite{tokmakov2021learning} addresses the limitations of CenterTrack in recovering objects after occlusions. 
        It assumes object permanence under occlusions and continues modeling the spatio-temporal movement of lost objects.
        LGM \cite{wang2021track} proposes a motion-based model for the vehicle tracking task, leveraging both local and global motion consistencies to track and recover vehicles after occlusions. 
        However, its applicability is limited to tracking vehicles with linear motion, lacking robustness in handling objects with non-linear motion.

        Along with these works, the SORT-series trackers \cite{bewley2016simple, wojke2017simple, zhang2022bytetrack, cao2023observation} utilize the Bayesian estimation \cite{lehmann2006theory} as a motion model.
        For instance, SORT \cite{bewley2016simple} employs the original Kalman Filter \cite{kalman1960contributions} with linear assumptions for object motion estimation and the Hungarian matching algorithm \cite{kuhn1955hungarian} to match predictions and detections.
        However, as motion features alone offer limited information, Deep-SORT \cite{wojke2017simple} and ByteTrack \cite{zhang2022bytetrack} propose a hybrid method by incorporating the visual features with the Kalman Filter predictions to enhance object discrimination.
        On the other hand, OC-SORT \cite{cao2023observation} improves robustness in handling occlusions without appearance information by prioritizing observations instead of linear estimations, but still struggles in recovering lost objects under non-linear motion and long-term occlusions.

    \subsection{Transformers in Multi-Object Tracking}\label{ssec:transformer_rel}

        In scenarios involving non-linear motion and occlusions, transformers demonstrate promising results for their inherent power to model complex interactions and adaptively process sequential information.
        As shown in Fig. \ref{fig:rel_work}(a), the existing transformer-based MOT approaches learn object queries to capture mainly appearance information \cite{sun2020transtrack, meinhardt2022trackformer, zeng2022motr, zhou2022global, gao2023memotr}.
        In particular, TransTrack \cite{sun2020transtrack} utilizes the transformer to extract the object-level appearance features and learn the aggregated visual embedding of each object for subsequent IoU-based matching.
        Since appearance information is sensitive to occlusions, TrackFormer \cite{meinhardt2022trackformer}, MOTR \cite{zeng2022motr} and MeMOTR \cite{gao2023memotr} jointly model both motion and appearance by representing each object as an autoregressive track query and recurrently propagating them to associate with identical instances across subsequent frames. 

        In contrast, AM-SORT only leverages motion information, employing simple and lightweight bounding boxes.
        To the best of our knowledge, AM-SORT stands out as the first successful application of transformers in purely motion-based methods.
        We believe that AM-SORT will encourage further research on adaptable motion predictors.


    \begin{figure*}[t]
        \centering
        \includegraphics[scale=0.095]{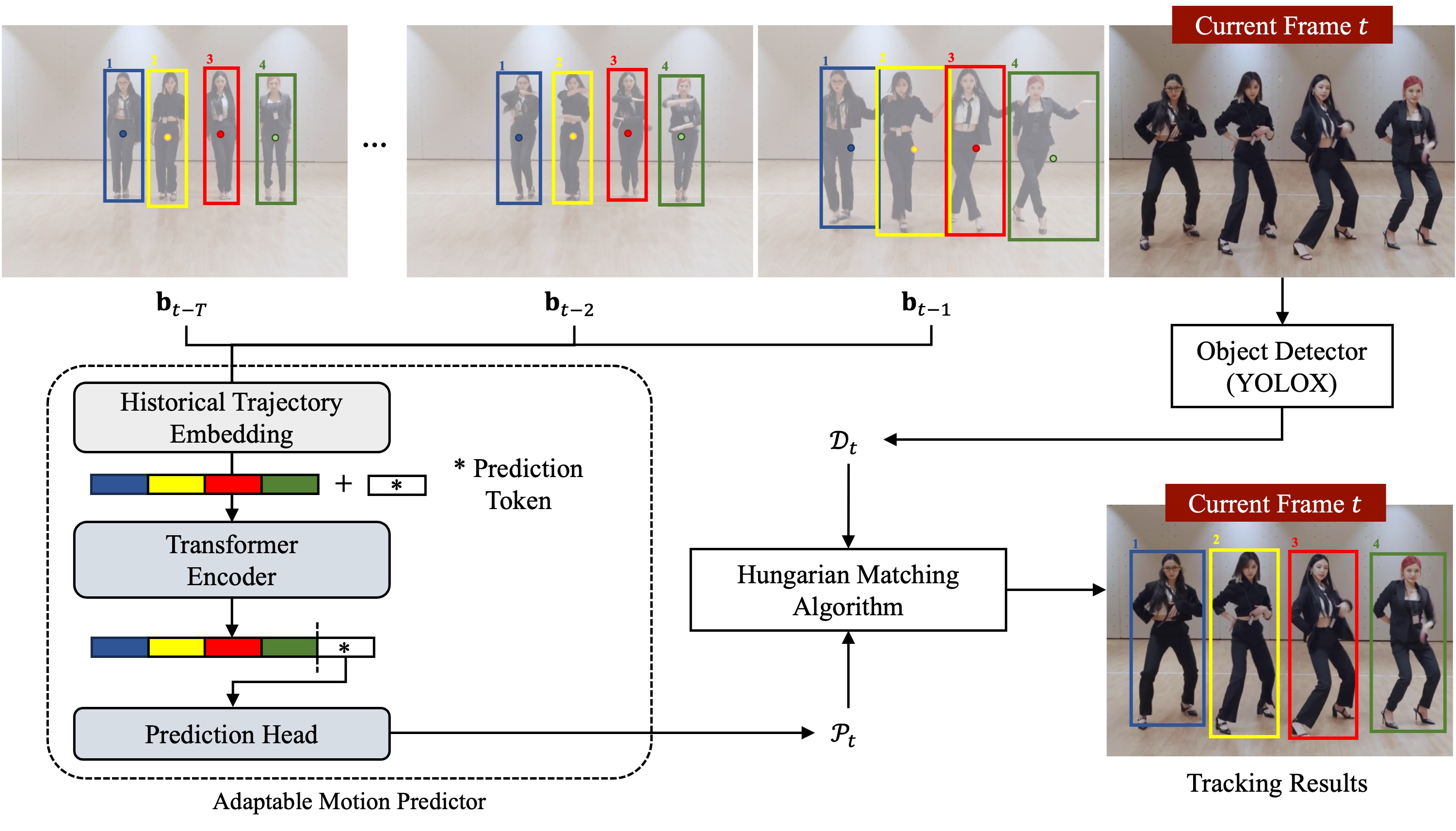}
        \vspace{-0.3cm}
        \caption{Illustration of the AM-SORT overall pipeline. The historical trajectory of length $T$ is fed into the transformer encoder to estimate the track predictions $\mathcal{P}_{t}$. Through utilizing an off-the-shelf detector, detections $\mathcal{D}_{t}$ are obtained. Subsequently, the Hungarian matching algorithm associates $\mathcal{D}_{t}$ with $\mathcal{P}_{t}$, resulting in the final output tracks.
        \vspace{-0.4cm}
        }
        \label{fig:overall_pipeline}
    \end{figure*}

\vspace{-0.2cm}
\section{Proposed Method}\label{sec:proposed_method}
    
    AM-SORT leverages motion cues to robustly track objects with non-linear motion patterns. 
    Our primary focus is on achieving accurate estimations of non-linear uncertainties by introducing an adaptable motion predictor based on the transformer encoder which supersedes the Kalman Filter. 
    Fig. \ref{fig:overall_pipeline} shows the overall pipeline of AM-SORT. 
    Specifically, we input the historical trajectory of an individual object containing a sequence of bounding boxes in the previous frames, denoted as 
    $\mathbf{B}_{t-T:t-1}=\{\mathbf{b}_{t-T},\ldots,\mathbf{b}_{t-2},\mathbf{b}_{t-1}\}$, 
    where $T$ is the pre-defined historical trajectory length. 
    The bounding boxes are represented as 
    $\mathbf{b}=(c_{x},c_{y},w,h)$, 
    where $(c_{x},c_{y})$ is the center coordinate of the object in the image plane, $w$ and $h$ stand for width and height, respectively. 
    The transformer encoder produces the refined prediction token, which is subsequently converted into a bounding box $\mathbf{\hat{b}}_{t}$ through the prediction head. 
    The estimated bounding boxes generate a set of track predictions for the current frame, denoted as $\mathcal{P}_{t}$. Subsequently, detections in the corresponding frame, referred to as $\mathcal{D}_{t}$, are associated with $\mathcal{P}_{t}$ based on Intersection-over-Union (IoU) using the Hungarian matching algorithm \cite{kuhn1955hungarian}.

        \begin{figure}[t]
            \centering
            \includegraphics[scale=0.13]{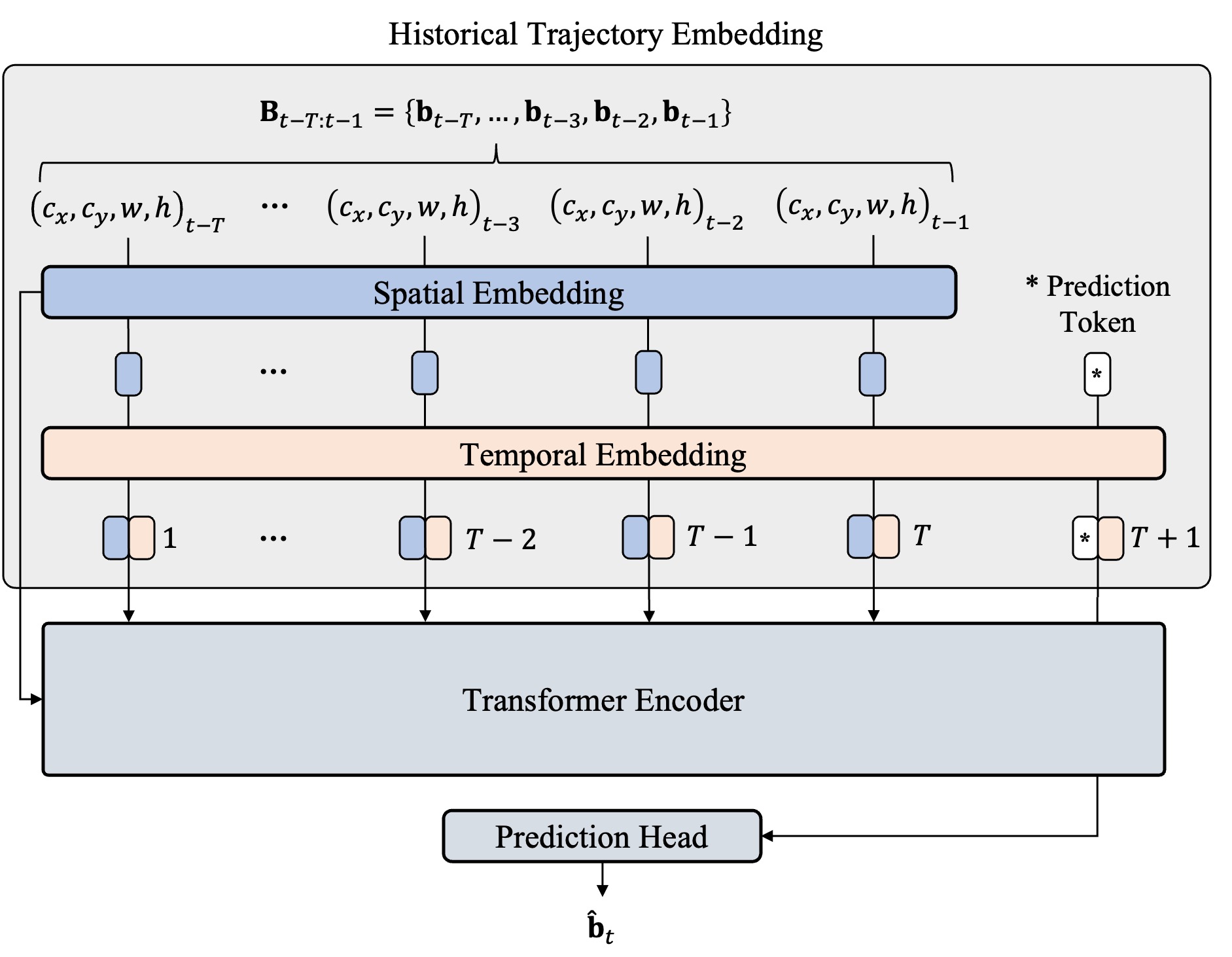}
            \vspace{-0.3cm}
            \caption{Illustration of our historical trajectory embedding in the motion predictor. The historical trajectory embedding encodes a comprehensive representation of a bounding box sequence by jointly considering spatio-temporal information.
            \vspace{-0.4cm}
            }
            \label{fig:traj_embed}
        \end{figure}

    \subsection{Historical Trajectory Embedding}
    \label{ssec:historical_trajectory_embedding}
        
        Historical trajectory embedding jointly encodes the spatial and temporal information from a sequence of bounding boxes and consists of three operations: spatial embedding, prediction token concatenation, and temporal embedding.
        Fig. \ref{fig:traj_embed} illustrates the structure of our historical trajectory embedding in the motion predictor.

        For spatial embedding, we utilize the sinusoidal positional encoding \cite{vaswani2017attention} to transform low-dimensional bounding boxes into a high-dimensional space to facilitate a fine-grained representation of each bounding box as follows:
        \begin{equation}
            \mathbf{x}_{t-T} = \mathrm{PE}_{\mathrm{spat}}(\mathbf{b}_{t-T}),
        \end{equation}
        where $\mathrm{PE}_\mathrm{spat}\!\!:\mathbb{R}^{4} \rightarrow \mathbb{R}^{D}$ represents the spatial embedding operation, $D$ is an embedding dimension and $\mathbf{x}_{t-T}$ denotes the spatial embedding of the bounding box.
        
        Subsequently, a prediction token is concatenated with the spatial embeddings at the end of the entire sequence. This prediction token is a learnable embedding that functions as a bounding box in the current frame $t$.
        The mathematical formulation is as follows:
        \vspace{-0.2 cm}
        \begin{equation}
            \mathbf{X}_{t-T:\mathrm{pred}} = \mathrm{Concat}( \mathbf{x}_{t-T}, \ldots, \mathbf{x}_{t-1} , \mathbf{x}_{\mathrm{pred}} ),
            \vspace{-0.2 cm}
        \end{equation}
        where $\mathbf{X}_{t-T:\mathrm{pred}}$ denotes the spatial embedding of the historical trajectory, obtained through the concatenation $\mathrm{Concat}(\cdot)$ of spatial embeddings and the prediction token $\mathbf{x}_{\mathrm{pred}}$. 

        For temporal embedding, we employ positional encoding similar to spatial embeddings.
        In contrast, we encode natural numbers which assign serial numbers to each spatial embedding in the sequence in reverse order from $T+1$ to $1$, starting from the last element.
        This ensures that the model prioritizes the terminal part of the historical trajectory embedding, even for objects with historical trajectory lengths less than $T$.
        Thus, the historical trajectory embedding is as follows:
        \vspace{-0.15 cm}
        \begin{equation}
            \mathbf{Z}_{t-T:\mathrm{pred}} = \mathbf{X}_{t-T:\mathrm{pred}} + \mathrm{PE}_{temp}(\mathbb{N}_{T+1:1}),
            \vspace{-0.15 cm}
        \end{equation}
        where $\mathbf{Z}_{t-T:\mathrm{pred}}$ represents our historical trajectory embedding, $\mathrm{PE}_{\mathrm{temp}}\!\!:\mathbb{R} \rightarrow \mathbb{R}^{D}$ denotes the temporal embedding and $\mathbb{N}_{T+1:1}$ is a sequence of natural numbers from $T+1$ to $1$.
        
        Notably, in the context of bounding box prediction where object localization is crucial, we enrich the historical trajectory embedding with additional spatial information before passing it through each encoder layer.

    \subsection{Adaptable Motion Predictor}
    \label{ssec:motion_pred}
        
        We utilize the transformer encoder as an adaptable motion predictor, which contains multi-head self-attention (MHSA) \cite{vaswani2017attention} layers and feed-forward neural networks. 
        MHSA facilitates interactions among each bounding box within the historical trajectory extracting their non-linear relationships. 
        This process refines the prediction token with sufficient information for precise localization of the object bounding box in the current frame, formulated as:
        \begin{equation}
            \mathbf{\hat{Z}}_{t-T:\mathrm{pred}} = \mathrm{Enc}(\mathbf{Z}_{t-T:\mathrm{pred}}),
        \end{equation}
        where $\mathrm{Enc}(\cdot)$ represents the transformer encoder operations, with $\mathbf{\hat{Z}}_{t-T:\mathrm{pred}}$ denoting the refined historical trajectory embedding.
        The prediction head receives only the prediction token $\mathbf{\hat{z}}_{\mathrm{pred}}$, which is the last element in the refined historical trajectory embedding, and utilizes it to generate the bounding box coordinates as follows: 
        \begin{equation}
            \mathbf{\hat{b}}_{t} = \mathrm{Head}(\mathbf{\hat{z}}_{\mathrm{pred}}),
        \end{equation}
        where $\mathrm{Head}(\cdot)$ denotes the prediction head and $\mathbf{\hat{b}}_{t}$ is the estimated bounding box in the current frame.
        The prediction head is composed of three linear layers each accompanied by a ReLU activation function, and the last layer utilizes a Sigmoid activation function to convert the bounding box coordinates in the range between 0 and 1.

    \subsection{Training}
    \label{ssec:training}

        We train our adaptable motion predictor by comparing the predicted bounding boxes with the ground truth. 
        We extract all the trajectories in an entire tracking video and segment them into bounding box sequences of length $T+1$. 
        The beginning bounding box sequence of each trajectory segment is utilized as a historical trajectory to estimate $\mathbf{\hat{b}}$ at the frame $T+1$, while the last bounding box $\mathbf{b}$ in the segment is considered as the ground truth. 
        We adopt the L1 loss function as the prediction loss to enhance robustness to outliers, such as errors in object detection and track prediction.
        Specifically, the estimated attributes $(\hat{c}_{x},\hat{c}_{y},\hat{w},\hat{h})$ of bounding box $\mathbf{\hat{b}}$ are compared to the respective attributes of ground truth $\mathbf{b}$ with L1 loss and our total prediction loss $\mathcal{L}_\mathrm{pred}$ is computed as the mean value:
            \vspace{-0.1cm}
            \begin{equation}
                \mathcal{L}_{\mathrm{pred}}(\mathbf{\hat{b}},\mathbf{b})=\frac{1}{4}\sum_{i}|\hat{b}_{i}-b_{i}|, \quad i\in{(c_{x},c_{y},w,h)}.
            \end{equation}

        \vspace{-0.6cm}
        \subsubsection{Masked Tokens.}
            
            We employ masked tokens as an augmentation strategy to simulate the effect of non-linear motion and occlusions.
            We mask bounding boxes within historical trajectories with a probability $p$.
            Subsequently, the masked bounding boxes are replaced by masked tokens to prevent the encoding of their spatial information.
            These masked tokens are represented as learnable embeddings, that are initialized with random values and optimized during training.
            In this manner, we enhance our model to gain a clear comprehension of missing trajectory segments.
            Our augmentation strategy with masked tokens facilitates effective masking operations, ensuring robust training in complex scenarios.
            
            Additionally, we utilize masked tokens to handle padding in historical trajectory embeddings during inference.
            We fill the historical trajectory embedding with masked tokens to maintain the constant length for newborn objects with past bounding boxes fewer than $T$.

\vspace{-0.2cm}
\section{Experiments}\label{sec:experiments}
    
    \subsection{Dataset and Evaluation Metric}\label{ssec:experiment_settings}
       
        We provide experimental results on DanceTrack \cite{sun2022dancetrack}, MOT17 \cite{milan2016mot16} and MOT20 \cite{dendorfer2020mot20}. 
        DanceTrack mainly consists of dance videos featuring objects with similar appearances. 
        DanceTrack provides scenarios characterized by non-linear object motion and occlusions, thereby posing significant challenges for motion-based tracking approaches.
        MOT17 and MOT20 contain pedestrian tracking videos in public spaces, where object motion is represented by slow and smooth movements, approximately linear. However, these datasets are still challenging due to highly crowded scenes with dense object populations.

        We use the evaluation metrics including HOTA (Higher Order Tracking Accuracy) \cite{luiten2021hota}, AssA (Association Accuracy) \cite{luiten2021hota}, DetA (Detection Accuracy) \cite{luiten2021hota}, IDF1 \cite{milan2016mot16} and  MOTA (Multi-Object Tracking Accuracy) \cite{milan2016mot16}. HOTA offers a balanced evaluation of both detection and association accuracy, in contrast to MOTA or DetA which is biased toward measuring detection. IDF1 and AssA are used to demonstrate the association performance.

    \vspace{-0.2cm}
    \subsection{Implementation Details}\label{ssec:implementation_details}

        We train our adaptable motion predictor on the corresponding tracking datasets without incorporating extra samples from other datasets.
        To ensure a fair comparison, we utilize the publicly accessible YOLOX \cite{ge2021yolox} detector weights developed by ByteTrack \cite{zhang2022bytetrack} for object detection following the baselines. 
        The transformer encoder is comprised of 6 layers with the multi-head self-attention employing 8 heads. 
        The embedding dimension $D$ is set to 512. 
        We use the Adam \cite{kingma2014adam} to optimize the network with a learning rate of 0.0001 for 50 epochs and set the batch size to 512.
        The historical trajectory embedding length $T$ is predefined as 30. 
        The masking probability $p$ is selected as 0.1.
        Analysis of the choice of $T$ and $p$ can be found in Section \ref{ssec:ablation_study}. 
        All experiments were conducted on a single NVIDIA TITAN XP.

        \begin{table*}[t]
            \centering
            \caption{Tracking results on the DanceTrack test set.}
            \begin{tabular}{l | c c | c c c c c}
            \toprule
            Tracker & \hspace{1mm} Appear. \hspace{0.5mm} & \hspace{0.5mm} Motion \hspace{1mm} & \hspace{1mm} HOTA$\uparrow$ & IDF1$\uparrow$ & MOTA$\uparrow$ & AssA$\uparrow$ & DetA$\uparrow$ \\ 
            \midrule
            DeepSORT \cite{wojke2017simple} & $\checkmark$ & $\checkmark$ & 45.6 & 47.9 & 87.8 & 29.7 & 71.0 \\
            ByteTrack \cite{zhang2022bytetrack} & $\checkmark$ & $\checkmark$ & 47.3 & 52.5 & 89.5 & 31.4 & 71.6 \\
            MOTR \cite{zeng2022motr} & $\checkmark$ & $\checkmark$ & 54.2 & 51.5 & 79.7 & 40.2 & 73.5 \\
            MeMOTR \cite{gao2023memotr} & $\checkmark$ & $\checkmark$ & 68.5 & 71.2 & 89.9 & 58.4 & 80.5 \\
            \midrule
            TransTrack \cite{sun2020transtrack} & $\checkmark$ & & 45.5 & 45.2 & 88.4 & 27.5 & 75.9 \\
            GTR \cite{zhou2022global} & $\checkmark$ & & 48.0 & 50.3 & 84.7 & 31.9 & 72.5 \\
            QDTrack \cite{fischer2023qdtrack} & $\checkmark$ & & 54.2 & 50.4 & 87.7 & 36.8 & 80.1 \\
            GHOST \cite{seidenschwarz2023simple} & $\checkmark$ & & 56.7 & 57.7 & 91.3 & 39.8 & 81.1 \\
            \midrule
            CenterTrack \cite{zhou2020tracking} & & $\checkmark$ & 41.8 & 35.7 & 86.8 & 22.6 & 78.1 \\
            TraDes \cite{wu2021track} & & $\checkmark$ & 43.3 & 41.2 & 86.2 & 25.4 & 74.5 \\
            SORT \cite{bewley2016simple} & & $\checkmark$ & 47.9 & 50.8 & \textbf{91.8} & 31.2 & 72.0 \\
            OC-SORT \cite{cao2023observation} & & $\checkmark$ & 54.6 & 54.6 & 89.6 & 40.2 & \textbf{80.4} \\
            AM-SORT (Ours) \hspace{1mm} & & $\checkmark$ & \textbf{55.6} & \textbf{56.3} & 89.6 & \textbf{40.4} & 80.3 \\
            \bottomrule
            \end{tabular}
            \label{tab:dancetrack}
            \vspace{-0.4cm}
        \end{table*}

    \vspace{-0.3cm}
    \subsection{Benchmark Results}
    \label{ssec:benchmark_results}
        Table \ref{tab:dancetrack} shows the benchmark results on the DanceTrack test set.
        AM-SORT achieves competitive performance compared to the appearance-based and hybrid trackers, and state-of-the-art results among motion-based MOT approaches. 
        It obtains 56.3 IDF1 and 55.6 HOTA, outperforming the baselines. 
        It is important to note that a significant gain of 1.7 is observed for IDF1, which measures association performance and re-identification accuracy. 
        
        Table \ref{tab:mot1720} shows the tracking performance on the MOT17 and MOT20 test sets to verify the generalizability covering linear object motion. AM-SORT achieves higher results compared to state-of-the-art MOT approaches. As mentioned earlier, MOT17 and MOT20 are designed for tracking pedestrians, where motion patterns are generally linear and do not contain non-linear scenarios. Despite these different conditions, AM-SORT still demonstrates consistent improvements, even though it does not align with primary issues.

        \begin{table}[t]
            \centering
            \caption{Tracking results on the MOT17 and MOT20 test sets under private detection protocols.}
            \begin{tabular}{l | c c c c | c c c c}
            \toprule
            \multicolumn{1}{c}{} & \multicolumn{4}{c}{MOT17} & \multicolumn{4}{c}{MOT20} \\
            \midrule
            Tracker & \hspace{0.5mm} HOTA$\uparrow$ & IDF1$\uparrow$ & MOTA$\uparrow$ & AssA$\uparrow$ & \hspace{0.5mm} HOTA$\uparrow$ & IDF1$\uparrow$ & MOTA$\uparrow$ & AssA$\uparrow$ \\ 
            \midrule
            \multicolumn{2}{l}{\textit{Hybrid:}} \\
            MOTR \cite{zeng2022motr} & 57.2 & 68.4 & 71.9 & 55.8 & 57.8 & 68.6 & 73.4 & / \\
            MeMOTR \cite{gao2023memotr} & 58.8 & 71.5 & 72.8 & 58.4 &  54.1 & 66.1 & 63.7 & 55.0 \\
            DeepSORT \cite{wojke2017simple} & 61.2 & 74.5 & 78.0 & 59.7 & 57.1 & 69.6 & 71.8 & 55.5 \\
            ByteTrack \cite{zhang2022bytetrack} & 63.1 & 77.3 & 80.3 & 62.0 & 61.3 & 75.2 & 77.8 & 59.9 \\
            \midrule
            \multicolumn{2}{l}{\textit{Appearance-based:}} \\
            QDTrack \cite{fischer2023qdtrack} & 53.9 & 66.3 & 68.7 & 52.7 & 60.0 & 73.8 & 74.7 & 58.9 \\
            TransTrack \cite{sun2020transtrack} & 54.1 & 63.9 & 74.5 & 47.9 & 48.9 & 59.4 & 65.0 & 45.2 \\
            GTR \cite{zhou2022global} & 59.1 & 71.5 & 75.3 & 57.0 & / & / & / & / \\
            GHOST \cite{seidenschwarz2023simple} & 62.8 & 77.1 & 78.7 & / & 61.2 & 75.2 & 73.7 & /  \\
            \midrule
            \multicolumn{2}{l}{\textit{Motion-based:}} \\
            SORT \cite{bewley2016simple} & 34.0 & 39.8 & 43.1 & 31.8 & 36.1 & 45.1 & 42.7 & 35.9 \\
            CenterTrack \cite{zhou2020tracking} & 52.2 & 64.7 & 67.8 & 51.0 & / & / & / & / \\
            TraDes \cite{wu2021track} & 52.7 & 63.9 & 69.1 & 50.8 &  / & / & / & / \\
            PermaTrack \cite{tokmakov2021learning} & 55.5 & 68.9 & 73.8 & 53.1 &  / & / & / & / \\
            OC-SORT \cite{cao2023observation} & 63.2 & 77.5 & \textbf{78.0} & 63.4 & \textbf{62.1} & 75.9 & \textbf{75.5} & \textbf{62.0} \\ 
            AM-SORT (Ours) \hspace{1mm} & \textbf{63.3} & \textbf{77.8} & \textbf{78.0} & \textbf{63.5} & 62.0 & \textbf{76.1} & \textbf{75.5} & 61.3 \\
            \bottomrule
            \end{tabular}
            \label{tab:mot1720}
            \vspace{-0.3cm}
        \end{table}

    \begin{figure}[t]
        \centering
        \includegraphics[width=\textwidth]{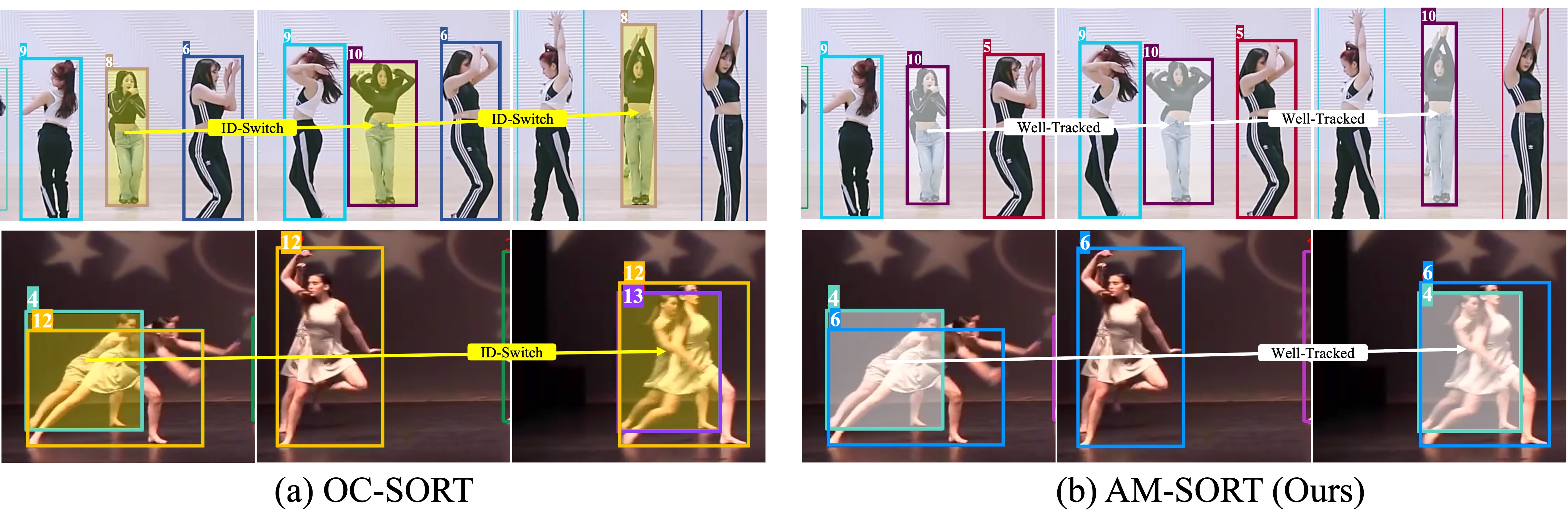}
        \vspace{-0.8cm}
        \caption{Qualitative comparison of OC-SORT and AM-SORT (Ours). The first row shows the tracking results in the scenario with non-linear changes of the bounding box for \textit{dancetrack0010} sequence; the second row in the scenario with non-linear object movement during occlusions for \textit{dancetrack0019}.
        }
        \label{fig:qual_comparison}
    \end{figure}

    \vspace{-0.2cm}
    \subsection{Qualitative Results}
    \label{ssec:qualitative}
        Fig. \ref{fig:qual_comparison} shows a qualitative comparison of OC-SORT and AM-SORT. These examples illustrate identity switches of the yellow-marked object in OC-SORT. 
        In Fig. \ref{fig:qual_comparison} Row 1, due to the linear assumptions inherent to the Kalman Filter, OC-SORT estimates a thin-shaped bounding box for the marked object in the middle frame. 
        It is unable to predict the sudden change in a wide-shaped bounding box leading to a false match. 
        Similarly, the linear assumptions prevent the capture of the directional shift to the right after occlusion in Fig. \ref{fig:qual_comparison} Row 2.
        In contrast, AM-SORT maintains consistent identities under these non-linear object motion and occlusions.

    \vspace{-0.2cm}
    \subsection{Ablation Study}
    \label{ssec:ablation_study}

        \subsubsection{Association Cost in Hungarian Matching Algorithm.}
            
            During inference, the SORT-series trackers utilize the Hungarian matching algorithm for object association.
            To show the impact of the association costs in the Hungarian matching step, we compare OC-SORT and AM-SORT at different combinations of association costs including IoU, motion direction difference $\Delta\theta$ and L1 distance. Motion direction difference calculates the direction similarity between existing tracks and new observations.
            AM-SORT with IoU alone as in Table \ref{tab:cost_matrix} Row 1 outperforms OC-SORT with IoU alone as in Table \ref{tab:cost_matrix} Row 1 by 3.2 IDF1 and achieves an increase of 0.2 IDF1 compared to OC-SORT with the best settings as in Table \ref{tab:cost_matrix} Row 2.
            On the other hand, motion direction difference degrades the tracking performance of our model. 
            The reason is that the motion direction cue, which is incorporated in OC-SORT to compensate for the approximately estimated bounding boxes in non-linear scenarios, is not suitable for AM-SORT.
            Our adaptable motion predictor already captures non-linear directional shifts in the prediction step making location-based matching sufficient.
            Furthermore, incorporating location-based association costs, IoU and L1 distance as in Table \ref{tab:cost_matrix} Row 3, gains an extra 2.5 IDF1 compared to OC-SORT with the best settings.

            \begin{table}[t]
                \centering
                \caption{Analysis at various association cost matrices.}
                \begin{tabular}{c c c | c | c}
                \toprule
                & & & \multicolumn{2}{c}{IDF1$\uparrow$} \\
                \midrule
                \hspace{2mm} IoU \hspace{2mm} & \hspace{2mm} $\Delta\theta$ \hspace{2mm} & \hspace{2mm} L1 \hspace{2mm} & \hspace*{4mm} OC-SORT \hspace*{4mm} & \hspace{2mm} AM-SORT (Ours) \hspace{2mm} \\
                \midrule
                $\checkmark$ &              &              & 51.1 & 54.3 \\
                $\checkmark$ & $\checkmark$ &              & \textbf{54.1} & 52.1 \\
                $\checkmark$ &              & $\checkmark$ & 53.4 & \textbf{56.5} \\
                $\checkmark$ & $\checkmark$ & $\checkmark$ & 53.2 & 53.2 \\
                \bottomrule
                \end{tabular}
                \label{tab:cost_matrix}
                \vspace{-0.4cm}
            \end{table}

        \vspace{-0.5cm}
        \subsubsection{Reliability of Bounding Box Predictions.}
            
            To verify the reliability of bounding box predictions, we evaluate OC-SORT and AM-SORT at progressively increased IoU thresholds. 
            The higher IoU threshold requires a larger overlap to associate detections with predictions.
            Table \ref{tab:iou_thresh} demonstrates that AM-SORT with IoU alone outperforms OC-SORT with the best settings at IoU thresholds greater than 0.4, while AM-SORT with the best settings achieves superior performance across all IoU threshold values.
            The higher IDF1 demonstrates that AM-SORT has a larger number of positively matched tracks with the ground truth. 
            This indicates that our adaptable motion predictor captures more accurately the object area, which serves as strong evidence for the higher reliability of the bounding box predictions.

            \begin{table}[t]
                \centering
                \caption{Analysis on prediction reliability at varying IoU thresholds.}
                \begin{tabular}{l | c c c c c c}
                \toprule
                & \multicolumn{6}{c}{IDF1$\uparrow$} \\
                \midrule
                IoU threshold & 0.3 & 0.4 & 0.5 & 0.6 & 0.7 & 0.8 \\
                \midrule
                OC-SORT & 54.1 & 51.5 & 46.9 & 38.7 & 25.5 & 15.8 \\
                AM-SORT (IoU) \hspace{1mm} & 54.3 & 51.2 & 48.4 & 40.4 & 27.0 & 16.4 \\
                AM-SORT (IoU+L1) & \hspace{1mm} \textbf{56.5} \hspace{1mm} & \hspace{1mm} \textbf{53.6} \hspace{1mm} & \hspace{1mm} \textbf{50.7} \hspace{1mm} & \hspace{1mm} \textbf{42.2} \hspace{1mm} & \hspace{1mm} \textbf{28.9} \hspace{1mm} & \hspace{1mm} \textbf{17.0} \hspace{1mm} \\
                \bottomrule
                \end{tabular}
                \label{tab:iou_thresh}
            \end{table}

        \vspace{-0.3cm}
        \subsubsection{Impact of Historical Trajectory Embedding Length.}
        
            To demonstrate how tracking performance varies with the historical trajectory embedding length, we evaluate AM-SORT at different $T$ values. 
            Table \ref{tab:traj_len} shows that performance increases with a longer historical trajectory and decreases when $T$ is greater than 30. 
            We conclude that a longer historical trajectory provides more comprehensive spatio-temporal information about object motion up to $T=30$.
            In contrast, overly old bounding boxes of the historical trajectory can provide noises and thus negatively impact the overall tracking performance.
            We set $T=30$, which covers 1.5 seconds of object trajectory in a 20 FPS video.

            \begin{table}[t!]
                \centering
                \caption{Impact of the historical trajectory embedding length $T$.}
                \begin{tabular}{c | c c c c c c}
                \toprule
                & \multicolumn{6}{c}{IDF1$\uparrow$} \\
                \midrule
                $T$ & 5 & 10 & 20 & 30 & 40 & 50 \\
                \midrule
                AM-SORT (Ours) \hspace{1mm} & \hspace{1mm} 52.9 \hspace{1mm} & \hspace{1mm} 53.5 \hspace{1mm} & \hspace{1mm} 55.2 \hspace{1mm} & \hspace{1mm} \textbf{56.5} \hspace{1mm} & \hspace{1mm} 55.0 \hspace{1mm} & \hspace{1mm} 54.3 \hspace{1mm} \\
                \bottomrule
                \end{tabular}
                \label{tab:traj_len}
            \end{table}

            \begin{table}[t!]
                \centering
                \caption{Impact of masked tokens at varying probabilities $p$.}
                \begin{tabular}{c | c c c c c c}
                \toprule
                & \multicolumn{6}{c}{IDF1$\uparrow$} \\
                \midrule
                $p$ & 0 & 0.05 & 0.1 & 0.2 & 0.3 & 0.4 \\
                \midrule
                AM-SORT (Ours) \hspace{1mm} & \hspace{1mm} 55.1 \hspace{1mm} & \hspace{1mm} 56.1 \hspace{1mm} & \hspace{1mm} \textbf{56.5} \hspace{1mm} & \hspace{1mm} 54.7 \hspace{1mm} & \hspace{1mm} 53.2 \hspace{1mm} & \hspace{1mm} 51.9 \hspace{1mm} \\
                \bottomrule
                \end{tabular}
                \label{tab:masking}
                \vspace{-0.4cm}
            \end{table}

        \vspace{-0.4cm}
        \subsubsection{Impact of Masked Tokens.}
        
            To show the effectiveness of utilizing masked tokens during training, we provide the tracking results for mask probabilities $p$ ranging from 0 to 0.4. 
            Table \ref{tab:masking} demonstrates that employing masked tokens with a probability of $p=0.1$ results in a 1.4 increase in IDF1 compared to training without masked tokens at $p=0$. 
            Conversely, masking with probability $p\ge0.2$ slightly drops the performance.
            We suggest that utilizing moderate masking of $p=0.1$ provides robust training to occlusions.

\vspace{-0.4cm}
\section{Conclusion}\label{sec:conclusion}
\vspace{-0.2cm}

    In this paper, we propose AM-SORT, a motion-based tracker with an adaptable motion predictor, that effectively addresses non-linear motion and occlusion. 
    We introduce historical trajectory embedding to encode spatio-temporal information in bounding box sequences for a comprehensive representation of object trajectory.
    We leverage the ability of transformers to model long-range dependencies in object trajectory, enabling our motion predictor to adapt to complex motion patterns.
    As a result, AM-SORT achieves competitive performance compared with state-of-the-art methods and outperforms existing motion-based approaches.

\vspace{-0.4cm}
\begin{credits}
\subsubsection{\ackname} This work was supported by the Institute of Information \& communications Technology Planning \& Evaluation (IITP) grant funded by the Korea government (MSIT) (No. 2019-0-00079, Artificial Intelligence Graduate School Program (Korea University)). Furthermore, we extend our sincere appreciation to Ho-Joong Kim for his invaluable support and feedback on the current research.

\end{credits}
\vspace{-0.4cm}
%
%
%
\bibliographystyle{splncs04}
\bibliography{main_paper}
%

\end{document}